\documentclass[11pt]{article}

\usepackage{emnlp2021}
\usepackage{times}
\usepackage{latexsym}
\usepackage[T1]{fontenc}
\usepackage[utf8]{inputenc}
\usepackage{microtype}
\usepackage{pgfplots}
\usepackage{xspace}
\usepackage{fdsymbol} 
\usepackage{booktabs}
\usepackage{makecell}
\usepackage{nth}
\usepackage{tabularx}
\usepackage{pifont}
\usepackage{multirow}
\usepackage{placeins}

\expandafter\def\expandafter\UrlBreaks\expandafter{\UrlBreaks
  \do\a\do\b\do\c\do\d\do\e\do\f\do\g\do\h\do\i\do\j%
  \do\k\do\l\do\m\do\n\do\o\do\p\do\q\do\r\do\s\do\t%
  \do\u\do\v\do\w\do\x\do\y\do\z\do\A\do\B\do\C\do\D%
  \do\E\do\F\do\G\do\H\do\I\do\J\do\K\do\L\do\M\do\N%
  \do\O\do\P\do\Q\do\R\do\S\do\T\do\U\do\V\do\W\do\X%
  \do\Y\do\Z}

\usepackage{soulutf8}
\usepackage[normalem]{ulem}
\usepackage{xcolor}

\usepackage{enumitem}
\setlist{nolistsep}
\usepackage{microtype}

\iftrue
  \usepackage[final]{proofing}
  \renewcommand\hl[1]{{#1}}  
   {\draftnote{\red{#2}}}
   \newcommand\redHL[1]{}
  \newcommand\todo[1]{}

  \newcommand{\Djame}[1]{}

\else  

\usepackage[draft]{proofing}

\newcommand{\Djame}[1]{
\textbf{\textcolor{red}{\hl{Djame: #1}}}
}

\newcommand\red[1]{{{\textcolor{red}{\bf #1}}}}

\let\oldred\red
\renewcommand\red[1]{{ \oldred{{#1}}}}

 \newcommand\redHL[1]{\red{\hl{#1}}}
\let\olddraftnote\draftnote
\renewcommand\draftnote[1]{\olddraftnote{\red{#1}}}

\fi

\iffalse
\author{José Carlos Rosales Núñez \\
Université Paris-Saclay \& CNRS, LISN \\
  91403 Orsay, France \\
  Inria Paris \\
  \texttt{jose.rosales@limsi.fr} \\\And
  Djamé Seddah \\
  Inria Paris \\
  F-75012, Paris, France \\
  \texttt{djame.seddah@inria.fr} \\\And
  Guillaume Wisniewski \\
  LLF, Université de Paris \& CNRS \\
  F-75013 Paris, France \\
  \texttt{guillaume.wisniewski@u-paris.fr} \\}

\else
\author{José Carlos Rosales Núñez$^{1,2,3}$ \quad Djam\'e Seddah$^3$ \quad Guillaume Wisniewski$^4$ \quad\\
$^1$ CNRS, LISN \quad $^2$ Universit\'e Paris-Saclay \\
$^3$ Inria Paris \quad $^4$ Université de Paris, LLF, CNRS\\
{\tt jose.rosales@limsi.fr} \quad {\tt djame.seddah@inria.fr} \\
  {\tt guillaume.wisniewski@univ-paris-diderot.fr} \\}

\fi

\title{Understanding the Impact of UGC Specificities on Translation Quality}
\date{}

\definecolor{mygrey}{RGB}{229,229,229}
\definecolor{mygrey2}{RGB}{127,127,127}
\definecolor{mygrey3}{RGB}{240,240,240}

\pgfplotsset{
 	axis background/.style={fill=mygrey},
	tick style=mygrey2,
	tick label style=mygrey2,
	grid=both,
	xtick pos=left,
	ytick pos=left,
	tick style={
		major grid style={style=white,line width=1pt},minor grid style=mygrey3,
		tick align=outside,
	},
	minor tick num=1,
}

\newcommand{\transformer}{\texttt{Transformer}\xspace}
\newcommand{\chartchar}{\texttt{char2char}\xspace}
\newcommand{\seqtseq}{\texttt{Seq2seq}\xspace}
\newcommand{\unk}{\texttt{<UNK>}\xspace}
\newcommand{\crapbank}{\texttt{PFSMB}\xspace}
\newcommand{\opensubs}{\texttt{OpenSubtitles}\xspace}
\newcommand{\blue}[1]{\textcolor{blue}{#1}}
\newcommand{\bleu}{\textsc{Bleu}\xspace}
\newcommand{\ourCorpus}{\texttt{PMUMT}\xspace}

\newcommand\review[3]{\textcolor{red}{\textbf{R#1.#2.}}\textcolor{blue}{#3}}
\iftrue
\renewcommand\review[3]{#3}
\fi
\usepackage{adjustbox}
\usepackage{array}

\newcolumntype{R}[2]{%
    >{\adjustbox{angle=#1,lap=\width-(#2)}\bgroup}%
    l%
    <{\egroup}%
}
\newcommand*\rot{\multicolumn{1}{R{45}{1em}}}

\begin{document}
\maketitle
\begin{abstract}
  This work takes a critical look at the evaluation of user-generated
  content automatic translation, the well-known specificities of which
  raise many challenges for MT. Our analyses show that measuring the
  average-case performance using a standard metric on a UGC test set
  falls far short of giving a reliable image of the UGC translation
  quality. That is why we introduce a new data set for the evaluation
  of UGC translation in which UGC specificities have been manually
  annotated using a fine-grained typology. Using this data set, we
 conduct several experiments to measure the impact of different kinds of UGC specificities
 on translation quality, more
 precisely than previously possible.

\end{abstract}

\section{Introduction}

This work takes a critical look at the evaluation of user-generated
content (UGC) automatic translation.  The well-known specificities of
UGC (high rate of OOVs, rare, grammatical constructs, ...) raise many
challenges for Machine Translation and has been the topic of many
recent
works~\cite{rosales-nunez-etal-2019-comparison,specia-etal-2020-findings}.

Several UGC parallel
corpora~\cite{michel-neubig-2018-mtnt,rosales-nunez-etal-2019-comparison}
have been introduced to evaluate the robustness of MT, some of which,
such as \cite{fujii-etal-2020-phemt}, are specially annotated to
identify UGC idiosyncrasies allowing to measure the impact of a given
specificity.  Our analyses (\textsection\ref{sec:exp}), indeed, show
that measuring the average-case performance using a standard metric on
a UGC test set falls far short of giving a reliable image of the UGC
translation quality: explaining the observed performance gap requires
a particular evaluation framework made of tailored metrics and specific
test sets in which UGC idiosyncrasies have been precisely annotated.

That is why, following this line of works, we introduce \ourCorpus, a
new parallel data set for the evaluation of UGC translation between
French and English in which UGC specificities have been manually
annotated using a fine-grained typology. \ourCorpus is larger, relies
on a more refined error typology and, more importantly, its
annotations are more detailed than existing noisy parallel corpora.
Its annotation scheme enables us to generate automatically parallel corpora in which
the kind and number of UGC specificities are precisely controlled. Contrary to many works studying the robustness of NMT systems by adding artificial noise to canonical corpora, \ourCorpus is made of attested UGC examples.

Using this framework, we conduct several experiments on three
out-of-the-box NMT architectures in a zero-shot scenario, to measure
more precisely than what was possible before the impact of the
different kinds of UGC specificities on translation
quality. Surprisingly enough, our experiments
(\textsection\ref{sec:analysis}) on natural data show that out-of-the-box models
exhibit unexpected strong robustness against several kinds of noise,
questioning several results reported in the
literature~\cite{michel-neubig-2018-mtnt,DBLP:conf/iclr/BelinkovB18}. We
believe that this data set and its associated evaluation framework
will pave the way for a better understanding of the interactions at
play in neural machine translation of noisy user-generated content
contexts.

\section{Testing Out-of-the-Box NMT models on UGC
  \label{sec:exp}}

\iftrue
\begin{table*}[h!]
{
{\footnotesize
  \begin{tabular}{cccccccc}
    \toprule
     $\downarrow$ Metric  / Test set $\rightarrow$ & \texttt{PFSMB}$^\dag$ & \texttt{PMUMT}$^\dag$ & \texttt{MTNT}$^\dag$ & \texttt{4SQ}$^\dag$ &\texttt{NewsTest} & \texttt{OpenSubsTest}  \\
    \midrule
    3-gram KL-Div & 1.563 & 1.442 & 0.471 & 0.500 & 0.406  & 0.006\\
    \%OOV  & 12.63 & 11.47& 6.78 & 3,46 & 3.81 & 0.76 \\
    PPL  &  599.48 & 596.12 & 318.24 & 293.67 & 288.83 & 62.06 \\
         \bottomrule
  \end{tabular}
}
}
  \centering
  \caption{Domain-related measure on the source side (FR), between
    used Test sets and other noisy UGC corpora using \opensubs as training set. Dags indicate
    UGC corpora.  {\em \texttt{4SQ} is the 4Square UGC data set introduced in \cite{4SQ}. PPL: perplexity,  KL-Div: Kullback-Leibler divergence.}  \label{tab:stat_test}}
\end{table*}

\fi

\begin{table*}[!htpb]
\footnotesize\centering
\begin{center}
\resizebox{\textwidth}{!}{\footnotesize
\begin{tabular}{lcccccccccc}
\toprule
   & \phantom{ab} & \multicolumn{4}{c}{\texttt{WMT}} & \phantom{abc} &\multicolumn{4}{c}{\texttt{OpenSubtitles}}  \\
  \cmidrule{3-6} \cmidrule{8-11}
   &  \phantom{ab} & \crapbank$^\dag$ & \texttt{MTNT}$^\dag$ & \texttt{News}$^\diamond$ &  \texttt{OpenTest} & & \crapbank$^\dag$ & \texttt{MTNT}$^\dag$ &  \texttt{News} &  \texttt{OpenTest}$^\diamond$   \\
  \midrule
  \multicolumn{11}{l}{\textit{BPE-based models}} \\
  \phantom{ab}\seqtseq           & \phantom{ab} & \phantom{0}9.9 & 21.8         & 27.5                & 14.7         & & 17.1 & 27.2 & 19.6 & 28.2 \\ 
  \phantom{abcd} + \unk rep. & \phantom{ab} & \textbf{17.1}  & \textbf{24.0}& \textbf{29.1} & \textbf{16.4} & & 26.1 & \textbf{28.5} & 24.5 & 28.2 \\ 
  \phantom{ab}\texttt{Transformer} & \phantom{ab} & 15.4 & 21.2 & 27.4 & \textbf{16.4} & & \textbf{27.5} & 28.3 & \textbf{26.7} & \textbf{31.4}  \\ 
  \midrule
  \multicolumn{11}{l}{\textit{Character-based models}} \\
  
  \phantom{ab}\chartchar & & \phantom{0}7.1 & 13.9 & 18.1 & \phantom{0}8.8  & & 23.8 & 25.7 &  17.8 & 26.3\\ 
    
  \bottomrule
\end{tabular}%
}
\caption{\textsc{Bleu} scores for our models. The $\dag$ symbol indicates the UGC test sets, and $\diamond$ in-domain test sets.
  \label{tab:res_bleu}}
\end{center}
\vspace{-1.5em}
\end{table*}

\subsection{Experimental Setting \label{sec:exp_setting}}

\paragraph{Training Data}
Because of the lack of a large parallel data set of noisy sentences,
we train our systems on `standard' parallel data sets:
\texttt{WMT}~\cite{bojar16findings} and
\opensubs~\cite{lison18opensubtitles2018}. The former contains
canonical texts (2.2M~sent.) and the latter (9.2M sent.) is made of
informal dialogues found in popular sitcoms.  

\paragraph{UGC Test Sets} To evaluate the different NMT models, we
consider two data sets of manually translated UGC:
\texttt{MTNT}~\cite{michel-neubig-2018-mtnt}  and the
Parallel French Social Media Bank corpus (\crapbank)~\cite{rosales-nunez-etal-2019-comparison}\footnote{\url{https://gitlab.inria.fr/seddah/parallel-french-social-mediabank}} which extends the French Social Media Bank \cite{seddahetal:2012:crapbank} with English translations. These two data sets 
raise many challenges for MT systems: they notably contain
characters that have not been seen in the training data (e.g.\
emojis), rare character sequences (e.g.\ inconsistent casing or
usernames) as well as many OOVs denoting URL, mentions,
hashtags or more generally named entities (NE). Most of the time,
OOVs are exactly the same in the source and target sentences.

\paragraph{NMT Models \label{sec:mt_models}\footnote{Models
    parameters are detailed in the appendix.}}

In our experiments, we use three translation models.  The
first two models are standard NMT models that take as input BPE
tokenized sentences: the model used in \citep{michel-neubig-2018-mtnt}, a
\seqtseq bi-LSTM architecture with global attention decoding as
implemented in \texttt{XNMT}~\cite{DBLP:conf/amta/NeubigSWFMPQSAG18}
as well as a vanilla \transformer model as implemented in the
\texttt{OpenNMT} toolkit~\cite{DBLP:conf/amta/KleinKDNSR18}.

We also consider a char-based model, namely the \chartchar of
\newcite{DBLP:journals/tacl/LeeCH17_et_al}. Using char-based models
which are, by nature, open-vocabulary to translate UGC is intuitively
appealing as these models are designed specifically to address the
problem of translating OOVs and to deal with noisy
input~\cite{DBLP:conf/iclr/BelinkovB18}.

%

As the \seqtseq model we consider in our experiments is not able to
translate OOVs, we introduce, as part of our translation
pipeline, a post-processing step in which the translation hypothesis
is aligned with the source and \unk
tokens are replaced by their aligned source
token. In the case of our \transformer model, \textsc{OpenNMT} performs this automatically.

\subsection{Results}

Table \ref{tab:res_bleu} reports the \bleu scores \cite{DBLP:conf/acl/PapineniRWZ02}
\footnote{All \bleu
  scores are calculated by \newcite{post18call}'s \texttt{SacreBleu}
  using the \textit{intl} tokenization} of the different models we
consider both on canonical and non-canonical test sets. Contrary to
the first results of \newcite{michel-neubig-2018-mtnt}, the quality of UGC
translation does not appear to be so bad: the drop in performance
observed on non-canonical corpora is of the same order of magnitude as
the drop observed when translating out-of-domain data.

These results seem to indicate that, counter-intuitively, translating
UGC does not raise any specific challenges. We however believe that
they are biased by the evaluation metric used: as UGC contains many
mentions, URLs emoticons, or named entities that are the same in the
source and in the target sentence, \bleu scores estimated on a
canonical and on a non-canonical can not be directly compared: \bleu
scores on non-canonical data are artificially high as systems are
rewarded for simply coping source tokens, which is the most natural
solution to translate OOVs. For instance, the \bleu score between the
\emph{sources} and references of the \crapbank is 15.1 while it is
only 2.7 on the \texttt{WMT} test set.
That is why, we believe that the usual MT metrics overestimate the
translation quality on UGC and we introduce, in the next section, a
new corpus and a new way to measure the real impact of UGC
specificities on translation quality.

\section{Analyzing the Impact of UGC on Translation Quality \label{sec:analysis}}

\begin{table*}[htbp]
\centering
  {\footnotesize
    \begin{tabular}{rccccccccccccc}
      & \rot{char del/add}
      & \rot{missing diacritics}
      & \rot{phonetic writing}
      & \rot{tokenization}
      & \rot{wrong tense}
      & \rot{special char}
      & \rot{agreement}
      & \rot{casing}
      & \rot{emoji}
      & \rot{named entity}
      & \rot{contraction}
      & \rot{repetition}
      & \rot{interjections} \\
      \midrule 
      \multirow{2}{*}{s2s} &
            \makecell{0.80 \\ {\scriptsize(28.7)}} &
            \makecell{0.95 \\ {\scriptsize(33.9)}} & 
            \makecell{0.93 \\ {\scriptsize(27.3)}} & 
            \makecell{0.96 \\ {\scriptsize(30.8)}} &
            \makecell{0.94 \\ {\scriptsize(30.7)}} & 
            \makecell{0.88 \\ {\scriptsize(26.1)}} & 
            \makecell{0.95 \\ {\scriptsize(27.1)}} & 
            \makecell{0.75 \\ {\scriptsize(27.7)}} & 
            \makecell{0.91 \\ {\scriptsize(31.0)}} & 
            \makecell{0.86 \\ {\scriptsize(31.7)}} &
            \makecell{0.95 \\ {\scriptsize(30.8)}} &
            \makecell{0.90 \\ {\scriptsize(30.2)}} &
            \makecell{0.93 \\ {\scriptsize(29.2)}} \\ 
     \midrule
         \multirow{2}{*}{c2c}  &      
            \makecell{0.99 \\ {\scriptsize(32.5)}} &
            \makecell{0.99 \\ {\scriptsize(29.6)}} & 
            \makecell{0.86 \\ {\scriptsize(25.2)}} & 
            \makecell{1.00 \\ {\scriptsize(31.9)}} &
            \makecell{0.97 \\ {\scriptsize(28.8)}} & 
            \makecell{0.81 \\ {\scriptsize(24.6)}} & 
            \makecell{0.96 \\ {\scriptsize(28.9)}} & 
            \makecell{0.86 \\ {\scriptsize(28.0)}} & 
            \makecell{0.83 \\ {\scriptsize(26.2)}} & 
            \makecell{0.94 \\ {\scriptsize(32.7)}} &
            \makecell{0.91 \\ {\scriptsize(30.4)}} &
            \makecell{0.95 \\ {\scriptsize(26.2)}} &
            \makecell{0.91 \\ {\scriptsize(28.7)}} \\ 
            
            \midrule 
            \multirow{2}{*}{TX}        &
            \makecell{0.98 \\ {\scriptsize(35.3)}} &
            \makecell{1.02 \\ {\scriptsize(34.0)}} & 
            \makecell{1.03 \\ {\scriptsize(33.2)}} & 
            \makecell{0.98 \\ {\scriptsize(32.9)}} &
            \makecell{1.02 \\ {\scriptsize(33.7)}} & 
            \makecell{0.92 \\ {\scriptsize(29.2)}} & 
            \makecell{0.97 \\ {\scriptsize(33.8)}} & 
            \makecell{0.90 \\ {\scriptsize(26.9)}} & 
            \makecell{0.75 \\ {\scriptsize(28.3)}} & 
            \makecell{0.99 \\ {\scriptsize(35.4)}} &
            \makecell{0.93 \\ {\scriptsize(31.1)}} &
            \makecell{0.89 \\ {\scriptsize(36.8)}} &
            \makecell{0.86 \\ {\scriptsize(30.2)}} \\         
         \bottomrule \end{tabular} \caption
    {\bleu score ratios between pairs of noisy and normalized
      sets of sentences, containing only one UGC
      specificity. \bleu scores on noisy sets are shown in
      parenthesis. 
      \label{tab:BLEU_ratio_1error}}
}
\end{table*}

In order to understand the impact of UGC specificities on translation
quality, we have annotated a new corpus in which UGC peculiarities are
identified in each source sentence and `normalized' to a canonical
form. 

\subsection{A Corpus Annotated with UGC Specificities \label{sec:corpus}}

\paragraph{The \ourCorpus corpus} To understand the impact
of UGC peculiarities, we manually annotated 400 source sentences
sampled from the \crapbank: \review{2}{2 - ok for me -ds}{one of the authors, fluent in French and with
good knowledge of UGC,} has identified spans in the sentence that differ
from canonical French and characterized these specificities using the
fine-grained typology of \newcite{DBLP:journals/corr/abs-2011-02063}
(see Table~\ref{tab:error_types}). \review{1-2}{3-2 ok for me}{Since the whole 
annotation process was done by a single person, no inter-annotator 
agreement can be calculated. Nevertheless,  results of our pilot 
study for each individual UGC peculiarity (cf. Table~\ref{tab:BLEU_ratio_1error} and  Table~\ref{tab:detailled_BLEU_ratio_1error} in the Appendix, for a cross-metrics analysis), show that  MT performance consistently performs better on our normalized corpus than on the original noisy set.}

\begin{table}[!htpb]
  \footnotesize
  \begin{tabular}{ll}
    \toprule
    code & kind of specificities \\
    \midrule
    1 & Letter deletion/addition\\
    2 & Missing diacritics \\
    3 & Phonetic writing\\
    4 & Tokenisation error\\
    5 & Wrong verb tense\\
    6 & \#; @, URL\\
    7 & Wrong gender/grammatical number\\
    8 & Inconsistent casing\\
    9 & Emoji\\
    10 & Named Entity\\
    11 & Contraction\\
    12 & Graphemic/punctuation stretching\\
    13 & Interjections\\
    \bottomrule
  \end{tabular}
  \centering
  \caption{Typology of UGC specificities used in our manual
    annotation.\label{tab:error_types}}
\end{table}

Each span containing an UGC specificity has been `normalized' to a
form closer to canonical French.\footnote{To ensure that this
  normalization has actually made our corpus closer to a canonical
  corpus, we have computed the perplexity of the original sentences
  and of the normalized sentences estimated by a 5-gram Kneser-Ney
  language model trained on the \opensubs corpus: the normalized
  version has a perplexity of 2,214 (and 11.60\% of its token are OOVs)
  far lower that the original version (with a perplexity of 8,546 and
  an OOV ratio of 19.60\%).} Table~\ref{tab:annotated_examples} shows
  some examples of annotated (source) sentence.
%
%
  A normalized form of each target (i.e.\ English) sentence has also
  been produced to ensures that the target can be generated from the
  `normalized' source.
  
  In the end, the annotation of this corpus
  represents 200h of work, comprising an iterative improvement 
  and debugging of the annotations to achieve the corpus' current version.
  \footnote{The annotated corpus and code collection can be found
    in \\\url{https://github.com/josecar25/PMUMT_annotated_UGC_corpus/}}

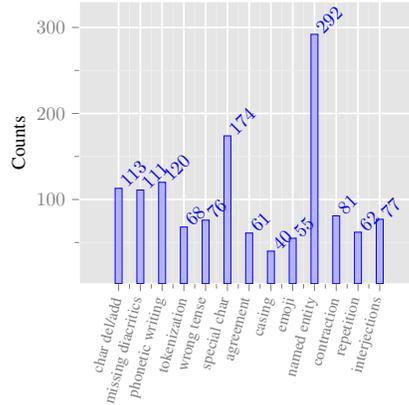
\begin{figure}[!htpb]
\centering
\begin{tikzpicture}[scale=0.7]
\begin{axis}[
    draw=white,
    ybar=0pt,
    bar width=4pt,
    enlargelimits=0.15,
    legend style={at={(0,1)},anchor=north west},
    ylabel={Counts},
    x tick label style={font=\footnotesize,rotate=75,anchor=east,},
    xticklabels={
      {char del/add},
      {missing diacritics},
      {phonetic writing},
      {tokenization},
      {wrong tense},
      {special char},
      {agreement},
      {casing},
      {emoji},
      {named entity},
      {contraction},
      {repetition},
      {interjections}},
    xtick=data,
    nodes near coords,
    nodes near coords align={vertical},
    every node near coord/.append style={rotate=45, anchor=west},
    ]
\addplot coordinates {(1,113) 
                      (2,111) 
                      (3,120) 
                      (4,68) 
                      (5,76) 
                      (6,174) 
                      (7,61) 
                      (8,40) 
                      (9,55) 
                      (10,292) 
                      (11,81) 
                      (12,62)
                      (13,77)};
\end{axis}
\end{tikzpicture}
\caption{\label{fig:type_distribution} Distribution of UGC specificites of the FR UGC sources in \ourCorpus.}
\vspace{-1.5em}
\end{figure}

The resulting corpus contains more
than $1,310$ annotations. On average, each sentence contains $2.8$ UGC
peculiarities. 
Figure~\ref{fig:type_distribution} describes the
distribution of UGC peculiarities in the corpus. 
%
  
\paragraph{Controlling the Number of Specificities per Sentence} 
Comparing the
predictions of an NLP system taking either the normalized sentences or
the original non-canonical sentences as input allows us to measure the
impact of UGC on this system. However, it is impossible to perform a
fine-grained analysis in which, for example, the impact of different
types of specificities are compared, since UGC sentences generally
contain several specificities of different types and the interactions
between them cannot be easily neutralized
.

That is why we have also constructed automatically a second version
of our corpus to help us analyze the interactions between the UGC
specificities in a sentence: by substituting only some of the span we
have annotated, we can create corpora in which the number and the kind
of specificities present in each sentence is tightly controlled
. In this framework, each original sentence can be (partially)
rewritten into as many sentences as there are UGC specificities in it.

This possibility of partial substitution greatly reduces the amount of
data to be annotated for our analyses: instead of having to annotate a
large amount of data to find enough sentences fulfilling the requested
criteria, we are able to generate these sentences from our original
annotation of 400 sentences. 
We believe that this approach could be of great
interest to perform fine-grained error analysis for NLP systems
dealing with UGC.

\subsection{Impact of UGC Peculiarities on Translation Quality \label{sec:impact}}

We used the \ourCorpus corpus to evaluate the impact of UGC
peculiarities on translation quality: we have reported in
Table~\ref{tab:contrast} the \bleu scores achieved by the considered
systems on both the 400 original sentences and the 400 normalized
sentences.  As expected, translations of normalized sentences, that
are more similar to the training data, are of better quality than
translations of original (noisy) sentences: the \bleu scores achieved
when translating normalized UGC content are close to those obtained on
the in-domain test-set.

For all systems, considering the non-canonical original sentences
results in a drop in translation quality of the same order of
magnitude, which shows that, even if these models build sentence
representations from completely different information, the presence of
UGC peculiarities has a similar impact on all of them.

\begin{table}
  \begin{tabular}{lrr}
    \toprule
                         & original & normalized \\
                         \midrule
                         
    \seqtseq             & 25.8     & 32.4       \\
    \chartchar           & 24.1     & 30.5       \\
    \transformer         & 28.6     & 33.6       \\
    \bottomrule
  \end{tabular}
  \centering
  \caption{\bleu scores on the original and normalized source
    sentences of the \ourCorpus corpus. 
    \label{tab:contrast}}
    \vspace{-1.5em}
\end{table}

\paragraph{Individual UGC Errors} To get a more precise picture of the
impact of UGC on translation quality, we have computed, for each kind
of peculiarities, the \bleu scores achieved on the corpus built to
contain only this peculiarity and the \bleu score computed on the
`normalized' version of the same
sentences. Table~\ref{tab:BLEU_ratio_1error} reports the ratio between
these two scores (detailed results are reported in
Table~\ref{tab:detailled_BLEU_ratio_1error} in the supplementary
material).




The impact of a given kind of UGC specificity on translation quality
is very different from one system to another: it appears that the
source sentences representation that MT systems learn to construct are
not sensitive to the same kind of noise or errors in the source
sentence and even seem to be complementary. For instance, inconsistent
casing strongly penalizes the \seqtseq model but has only a limited
impact on the \chartchar model
. On the contrary, the presence of characters specific to
online conversation such as @ or \# results in a substantial decrease
of translation quality for \chartchar
, but has less impact for \seqtseq or \transformer
, suggesting that
char-based models are not able to properly modeled characters that
hardly appear in the training set.

Interestingly, the \transformer model appears to be very robust to a
wide array of UGC peculiarities, even if it was not designed
specifically to handle noisy inputs: in particular, the presence of
named entities, spelling errors (i.e. substitution, deletion or
insertion of letters), agreement error (of verb tense or in gender and
number) as well as \texttt{tokenization} errors hardly hurt
translation quality. 
Similarly, the \chartchar model succeed in translating correctly
sentences with letter addition or suppression, showing that the model
actually manage to learn sentence representations that are robust to
spelling errors even if such errors are not present at training
time. This results is at odd with the conclusion drawn by
\newcite{DBLP:conf/iclr/BelinkovB18} on artificial
data.

%


\paragraph{Combination of Peculiarities} 

To better understand the impact of combinations of UGC peculiarities
on translation quality, Table~\ref{tab:BLEU_ratio_Nerror} reports the
ratio between the \bleu scores computed on the translation of a corpus
in which there are exactly $N$ different UGC peculiarities in a
sentence and on the translation of the normalized version of these
sentences. It appears that for all our systems
translation quality decreases linearly with the number of
specificities, suggesting that the impacts of the different
specificities are independent of each other. Surprisingly enough, the
gap between the \chartchar and \transformer is getting
smaller with the number of specificities in each sentence.


\begin{table}[!htbp]
\centering
{\footnotesize
\begin{tabular}{lccccc}
       & 1 & 2 & 3 & 4+ \\ \midrule
  \# sents. & 1,306 & 1,776 & 1,439 & 1,089 \\ \midrule
  
  s2s  & \makecell{0.90 \\ {\scriptsize(30.1)}}
       & \makecell{0.83 \\ {\scriptsize(27.0)}}
       & \makecell{0.77 \\ {\scriptsize(24.2)}}
       & \makecell{0.75 \\ {\scriptsize(23.2)}} \\ \midrule

  c2c  & \makecell{0.92 \\ {\scriptsize(29.5)}} 
       & \makecell{0.87 \\ {\scriptsize(26.6)}} 
       & \makecell{0.83 \\ {\scriptsize(24.3)}} 
       & \makecell{0.83 \\ {\scriptsize(23.2)}} \\ \midrule
  
  TX   & \makecell{0.96 \\ {\scriptsize(32.8)}} 
       & \makecell{0.89 \\ {\scriptsize(30.0)}} 
       & \makecell{0.86 \\ {\scriptsize(28.3)}} 
       & \makecell{0.84 \\ {\scriptsize(26.5)}}\\ \bottomrule
\end{tabular}
}
\caption{\bleu score ratio between pairs of normalized and noisy
  sentences containing $N$ specificities. \bleu scores on noisy
  sentences are shown in parenthesis.
  \label{tab:BLEU_ratio_Nerror}}
\end{table}

\section{Conclusions}

This work introduces \ourCorpus a new corpus of UGC translation in
which UGC specificities are manually annotated using a fine-grained
typology. Thanks to our detailed annotation process, we were able to build a
new framework that allows us to automatically generate several
parallel corpora in which the number and kind of UGC specificities is
precisely controlled.

Our experiments show that, contrary to what was previously believed,
out-of-the-box NMT models are robust to many different kind of UGC specificities and that the different architectures we tested are complementary, in the sense that they are not sensitive to the same specificities.  In
our future work, we plan to explore the intricacies of the robustness that seem to be linked to specific UGC idiosyncrasies. We make this data set and its associated evaluation framework public\footnote{\url{https://github.com/josecar25/PMUMT_annotated_UGC_corpus/}} as we believe it can pave the way for a better understanding of the interactions at play in neural machine translation of noisy user-generated content contexts.

\section*{Acknowledgments}
We thank the reviewers for their valuable feedback. This work was funded by the French Research Agency via the ANR ParSiTi project (ANR-16-CE33-0021).

\bibliographystyle{acl_natbib}
\bibliography{anthology,acl2021}

\onecolumn
\appendix
\setcounter{figure}{0} \renewcommand{\thefigure}{A.\arabic{figure}} 
\setcounter{table}{0} \renewcommand{\thetable}{A.\arabic{table}} 

\pagebreak
\section*{Supplementary Materials}

\FloatBarrier

\begin{table*}[!htpb]
{\footnotesize

  \begin{tabularx}{\textwidth}{llX}
    \toprule
    \ding{192} & src       & \textbf{JohnDoe389 \blue{(10)}} qui n'arrive pas \textbf{a \blue{(2)} depasser \blue{(2)}} 1 a \textbf{\blue{(2)}} \textbf{FlappyBird \blue{(10)}} ... \textbf{ptddddr \blue{(12,13)}}\\ 
      & ref              & JohnDoe389 who can't score more than 1 at FlappyBird ... lmaooooo \\
      \cmidrule{2-3}
      & N. src               &  Jean qui n'arrive pas à dépasser 1 à Jean ... \\
      & N. ref              & Jean who can't score more than 1 at Jean... \\
 
    \midrule
    \ding{193} & src        & \textbf{\#CaMeVénèreQuand \blue{(6)}} le matin \textbf{a \blue{(2)}} 7h on me parle alors que je suis pas encore \textbf{réveiller. \blue{(5)}} \\  
      & ref          & \#ItAnnoysMeWhen in the moring at 7 am someone talks to me although I didn't wake up yet.\\
       \cmidrule{2-3}
      & N. src           & le matin à 7h on me parle alors que je suis pas encore réveillé. \\
      & N. ref          & in the moring at 7 am someone talks to me although I didn't wake up yet.\\ 
      \midrule
\ding{194} & src              & vu sa \textbf{tete \blue{(2)}} \textbf{c \blue{(3)}} normal \textbf{kon \blue{(3)}} \textbf{est \blue{(3)}} \textbf{jms \blue{(11)}} \textbf{parler \blue{(5)}} d'elle ! \\
      & ref              & in light of her face it's normal no one ever spoke about her! \\   
       \cmidrule{2-3}
      & N. src               & vu sa tête c'est normal qu'on a jamais parlé d'elle ! \\
      & N. ref              & in light of her face it's normal no one ever spoke about her! \\
      \midrule
          \ding{195} & src & y a ma cousine qui \textbf{joue a \blue{(2)}} flappy bird \textbf{\blue{(10)}} \textbf{mdrrrrrrrrrrr \blue{(12, 13)} elle et plus nuuul \blue{(12,7)} que moi}\\
      & ref              & my cousin plays flappy bird loooooooooool she's more hopeless than me \\
       \cmidrule{2-3}
      & N. src               & y a ma cousine qui joue à Jean Jean elle et plus nulle que moi \\
      & N. ref              & my cousin plays Jean Jean she's more hopeless than me \\
      \bottomrule    
  \end{tabularx}
}
  \caption{Examples from our annotated noisy UGC corpus. Source sentences have been annotated with UGC
    specificities of Table~\ref{tab:error_types} 
    (in blue) according to their numerical code. For each example, the original source and reference (\textit{src} and \textit{ref}) and their corresponding normalized version (\textit{N. src} and \textit{N. ref}) are shown.
    \label{tab:annotated_examples}}
\end{table*}

\begin{table*}[!htpb]
{\footnotesize

  \begin{tabularx}{\linewidth}{llXlll}
    \toprule
   \multicolumn{6}{c}{Letter deletion/addition/change}  \\ \midrule \midrule[0.3pt]
    \ding{192} & src       & \multicolumn{4}{l}{j'arrive pas à \textbf{boir normalemen}}\\ 
     & norm              & \multicolumn{4}{l}{j'arrive pas à boire normalement} \\
      & ref              & \multicolumn{4}{l}{I can't \textbf{drink normally}} \\
      & s2s              & \multicolumn{4}{l}{I can't \textbf{drink normal.}} \\
      & c2c              & \multicolumn{4}{l}{I can't \textbf{drink normally.}}\\
      & Tx               & \multicolumn{4}{l}{I can't \textbf{drink normal men.}} \\

    \midrule
    \ding{193} & src        & \multicolumn{4}{l}{Je conseille à \textbf{toux} ceux qui ont l'esprit disons, un peu fermé de regarder sur les "Français d'origine contrôlée"}  \\  
         & norm              & \multicolumn{4}{l}{Je conseille à tous ceux qui ont l'esprit disons, un peu fermé de regarder sur les "Français d'origine contrôlée"}  \\
      & ref          & \multicolumn{4}{l}{I advise \textbf{everyone} with a, let's say a little narrow mind to watch about the "Français d'origine contrôlée"} \\
      & s2s           & \multicolumn{4}{l}{I suggest \textbf{cough} those who have minds say, a little closed to look at the "frances of controlled origins"} \\
      & c2c          & \multicolumn{4}{l}{I counsel \textbf{those} who have the mind, a little close to looking at the French original controlled original controlled.} \\ 
            & Tx               & \multicolumn{4}{l}{I advise \textbf{anyone} with a mind, say, a little closed to look at the controlled French.} \\

      \midrule
      
%

      \midrule   
          \ding{196} & src & \multicolumn{4}{l}{le côté suis \textbf{tro cool} au quotidien et je \textbf{relach} tout \textbf{quan} j'ai bu} \\
           & norm              & \multicolumn{4}{l}{les gens qui m'aiment me détestent quand j'ai bu} \\
      & ref              & \multicolumn{4}{l}{my side \textbf{very cool} in everyday life and \textbf{loosen} everything \textbf{when} I've been drinking} \\
      & s2s           & \multicolumn{4}{l}{I've been drinking all the time and I've been drinking everything \textbf{quan} I've been drinking.} \\
      & c2c              & \multicolumn{4}{l}{the side of the daily \textbf{cool} side and \textbf{relacing} everything \textbf{when} I've been drinking} \\
      & Tx               & \multicolumn{4}{l}{I'm the \textbf{cool} side. I'm the cool one.}\\
      
            \midrule   
   \multicolumn{6}{c}{Tokenization}  \\ \midrule \midrule[0.3pt]
          \ding{197} & src & \multicolumn{4}{l}{\textbf{J'sais pas vous}, mais de voir la joie des grands joueurs comme Zlatan, Motta, Verratti je trouve ça magnifique} \\
           & norm              & \multicolumn{4}{l}{Je sais pas vous, mais de voir la joie des grands joueurs comme Jean, Jean, Jean je trouve ça magnifique} \\
      & ref              & \multicolumn{4}{l}{\textbf{I don't know about you}, but seeing the joy of great players like Zlatan, Motta, Verratti I think it's wonderful} \\
      & s2s           & \multicolumn{4}{l}{\textbf{I don't know you}, but seeing the joy of the great players like Zlatan, Motta, Verratti, I think it's beautiful.}  \\
      & c2c              & \multicolumn{4}{l}{\textbf{I don't know about you}, but to see the joy of great players like Zlatan, Motta, Varratt, I think it's beautiful.} \\
      & Tx               & \multicolumn{4}{l}{\textbf{I don't know about you}, but seeing the joy of big players like Zlatan, Motta, Verratti, I think it's beautiful.} \\
            \midrule   
            
          \ding{198} & src &\multicolumn{4}{l}{pendant que vous me laissez en chien à l'atelier mon score de flappy bird fait que \textbf{d augmenter}} \\
          & norm              & \multicolumn{4}{l}{pendant que vous me laissez en chien à l'atelier mon score de Jean fait que d'augmenter} \\ 
      & ref              & \multicolumn{4}{l}{while you're bailing out on me at the workshop my flappy bird score \textbf{is just increasing}} \\
      & s2s           & \multicolumn{4}{l}{when you leave me as a dog when you leave me as a dog at the workshop.}  \\
      & c2c              & \multicolumn{4}{l}{while you leave me dog at the workshop my flappy bird score \textbf{is that increasing}} \\
      & Tx               & \multicolumn{4}{l}{while you leave me as a dog at the workshop my flappy bird score \textbf{is just up.}} \\
      
                  \midrule   
            
          \ding{199} & src & \multicolumn{4}{l}{il \textbf{ma} dit que c'était normal aussi et que ça allait redescendre,} \\
          & norm        &  \multicolumn{4}{l}{il m'a dit que c'était normal aussi et que ça allait redescendre,} \\
      & ref              & \multicolumn{4}{l}{\textbf{he told me} it was normal too and that it would come down,} \\
      & s2s           & \multicolumn{4}{l}{\textbf{He said it} was normal, too, and it was going to go down,} \\
      & c2c              & \multicolumn{4}{l}{\textbf{He told me it} was normal, too, and it was going back,} \\
      & Tx               & \multicolumn{4}{l}{\textbf{He told me it} was normal, too, and it was gonna come down,} \\
            \midrule   
   \multicolumn{6}{c}{Inconsistent casing}  \\ \midrule \midrule[0.3pt]
             \ding{200} & src & Jean \textbf{DANS VOS YEUX} & \ding{201} & src   & \textbf{JE VIENS DE VOIR} Jean \textbf{ET} Jean \textbf{JE PEUX PLUS} \\ 
             & norm & Jean dans vos yeux && norm &   Je viens de voir Jean je peux plus \\
             & ref & Jean \textbf{IN YOUR EYES} && ref              &  I JUST WATCHED Jean AND Jean CAN'T TAKE IT\\
      & s2s           & Jean \textbf{D in VOSY} && s2s           & I'm going to kill Jean and Jean I can't believe it. \\
      & c2c              & Jean \textbf{in your eyes} &&  c2c              & I'm here to see Jean And Jean I can no longer. \\
      & Tx               & Jean \textbf{in your eyes} && Tx               & I just saw Jean and Jean again. \\
               \midrule  
  
   \multicolumn{6}{c}{Domain-specific characters and emojis}  \\ \midrule \midrule[0.3pt]
        \ding{202} & src & \multicolumn{4}{l}{Avec mes magnifiques jumeaux Jean et Jean \textbf{@maxcarver @Charlie\_Carver $\color{red} \varheartsuit$}} \\
          & norm        &  \multicolumn{4}{l}{Avec mes magnifiques jumeaux Jean et Jean} \\
      & ref              & \multicolumn{4}{l}{With my wonderful twins Jean and Jean \textbf{@maxcarver @Charlie\_Carver $\color{red} \varheartsuit$}} \\
      & s2s           & \multicolumn{4}{l}{with my gorgeous Jean and Jean \textbf{@maxarver @Carlie\_Carver @Carlie\_Carver \#}} \\
      & c2c              & \multicolumn{4}{l}{with my beautiful Jean twins, Jean Jean and Jean \textbf{Charlier Charlier Carver.}} \\
      & Tx               & \multicolumn{4}{l}{with my beautiful twins Jean and Jean \textbf{imexcarver \@Charlie\_Charver @Charver}} \\
      \bottomrule    
  \end{tabularx}
}
  \caption{Examples from our noisy UGC corpus showing the \texttt{Transformer}, \texttt{char2char} and \seqtseq predictions. Present UGC
    specificities of Table~\ref{tab:error_types} 
    (in blue) are marked in bold.
    \label{tab:MT_out_examples}}
\end{table*}

\FloatBarrier

As some of the data sets contain as few as 40 sentences, we have also
computed the 95\% confidence interval for all \bleu scores in Table~\ref{tab:detailled_BLEU_ratio_1error} using the
bootstrapping method described in \newcite{koehn-2004-statistical}.
The width of all intervals is smaller than 0.30 for the \bleu scores
(roughly 1\% of the corresponding score) and than 0.006 for the ratios,
which shows that we can trustfully compare their values.

Similarly, we have included results for the \textsc{chrF} \cite{popovic-2015-chrf} and \textsc{Multi-bleu-detok.perl} evaluation metrics since \textsc{SacreBLEU} showed a ratio between performances on noisy and clean text versions, indicating that the noisy version could be easier to translate than its normalization. In this regard, we can see in Table~\ref{tab:detailled_BLEU_ratio_1error} that at least one of the three reported metrics gives a ratio value equal or smaller than $1.0$ within the 95\% confidence interval error (CI Err.), suggesting that our normalization introduce limited artificial noise, comparable to the difference between correlated evaluation metrics.

\begin{table*}[!htbp]
  {\footnotesize
    \begin{tabular}{rcccccccccccccc}
      & Metric
      & \rot{char del/add}
      & \rot{missing diacritics}
      & \rot{phonetic writing}
      & \rot{tokenization}
      & \rot{wrong tense}
      & \rot{special char}
      & \rot{agreement}
      & \rot{casing}
      & \rot{emoji}
      & \rot{named entity}
      & \rot{contraction}
      & \rot{repetition}
      & \rot{interjections} \\
      \midrule 
      \multirow{6}{*}{s2s} & MB &
            \makecell{0.78 \\ {\scriptsize(25.3)}} &
            \makecell{0.94 \\ {\scriptsize(31.1)}} & 
            \makecell{0.92 \\ {\scriptsize(23.2)}} & 
            \makecell{0.95 \\ {\scriptsize(30.6)}} &
            \makecell{0.98 \\ {\scriptsize(28.6)}} & 
            \makecell{0.83 \\ {\scriptsize(24.8)}} & 
            \makecell{0.95 \\ {\scriptsize(25.1)}} & 
            \makecell{0.77 \\ {\scriptsize(26.7)}} & 
            \makecell{0.87 \\ {\scriptsize(29.6)}} & 
            \makecell{0.87 \\ {\scriptsize(30.9)}} &
            \makecell{0.91 \\ {\scriptsize(28.4)}} &
            \makecell{0.83 \\ {\scriptsize(29.9)}} &
            \makecell{0.90 \\ {\scriptsize(26.5)}} \\ 
            \cmidrule{2-15}
       &chrF &
            \makecell{0.93 \\ {\scriptsize(46.7)}} &
            \makecell{0.97 \\ {\scriptsize(53.1)}} & 
            \makecell{0.89 \\ {\scriptsize(43.1)}} & 
            \makecell{0.95 \\ {\scriptsize(50.9)}} &
            \makecell{0.99 \\ {\scriptsize(50.6)}} & 
            \makecell{0.91 \\ {\scriptsize(44.3)}} & 
            \makecell{0.98 \\ {\scriptsize(49.3)}} & 
            \makecell{0.76 \\ {\scriptsize(40.2)}} & 
            \makecell{0.94 \\ {\scriptsize(51.1)}} & 
            \makecell{0.94 \\ {\scriptsize(51.7)}} &
            \makecell{0.93 \\ {\scriptsize(47.3)}} &
            \makecell{0.93 \\ {\scriptsize(47.3)}} &
            \makecell{0.96 \\ {\scriptsize(48.0)}} \\    
            \cmidrule{2-15}        
        &SB &
            \makecell{0.80 \\ {\scriptsize(28.7)}} &
            \makecell{0.95 \\ {\scriptsize(33.9)}} & 
            \makecell{0.93 \\ {\scriptsize(27.3)}} & 
            \makecell{0.96 \\ {\scriptsize(30.8))}} &
            \makecell{0.94 \\ {\scriptsize(30.7)}} & 
            \makecell{0.88 \\ {\scriptsize(26.1)}} & 
            \makecell{0.95 \\ {\scriptsize(27.1)}} & 
            \makecell{0.75 \\ {\scriptsize(27.7)}} & 
            \makecell{0.91 \\ {\scriptsize(31.0)}} & 
            \makecell{0.86 \\ {\scriptsize(31.7)}} &
            \makecell{0.95 \\ {\scriptsize(30.8)}} &
            \makecell{0.90 \\ {\scriptsize(30.2)}} &
            \makecell{0.93 \\ {\scriptsize(29.2)}} \\ 
     \midrule
         \multirow{6}{*}{c2c} &  MB & 
            \makecell{1.00 \\ {\scriptsize(29.5)}} & 
            \makecell{1.00 \\ {\scriptsize(27.4)}} & 
            \makecell{0.85 \\ {\scriptsize(22.5)}} &
            \makecell{0.99 \\ {\scriptsize(29.7)}} &
            \makecell{0.97 \\ {\scriptsize(26.9)}} &
            \makecell{0.80 \\ {\scriptsize(23.5)}} & 
            \makecell{0.97 \\ {\scriptsize(25.5)}} & 
            \makecell{0.91 \\ {\scriptsize(27.7)}} & 
            \makecell{0.83 \\ {\scriptsize(25.1)}} &
            \makecell{0.95 \\ {\scriptsize(31.7)}} & 
            \makecell{0.88 \\ {\scriptsize(26.6)}} &
            \makecell{0.93 \\ {\scriptsize(28.0)}} &
            \makecell{0.91 \\ {\scriptsize(25.7)}} \\ 
            \cmidrule{2-15}
                   &chrF &
            \makecell{0.99 \\ {\scriptsize(48.5)}} &
            \makecell{1.00 \\ {\scriptsize(50.6)}} & 
            \makecell{0.92 \\ {\scriptsize(44.8)}} & 
            \makecell{0.95 \\ {\scriptsize(50.1)}} &
            \makecell{0.99 \\ {\scriptsize(49.1)}} & 
            \makecell{0.84 \\ {\scriptsize(44.0)}} & 
            \makecell{0.98 \\ {\scriptsize(49.9)}} & 
            \makecell{0.78 \\ {\scriptsize(40.6)}} & 
            \makecell{0.93 \\ {\scriptsize(49.5)}} & 
            \makecell{0.95 \\ {\scriptsize(51.6)}} &
            \makecell{0.92 \\ {\scriptsize(48.8)}} &
            \makecell{0.90 \\ {\scriptsize(47.8)}} &
            \makecell{0.95 \\ {\scriptsize(49.7)}} \\    
            \cmidrule{2-15}        
        &SB &
            \makecell{0.99 \\ {\scriptsize(32.5)}} &
            \makecell{0.99 \\ {\scriptsize(29.6)}} & 
            \makecell{0.86 \\ {\scriptsize(25.2)}} & 
            \makecell{1.00 \\ {\scriptsize(31.9)}} &
            \makecell{0.97 \\ {\scriptsize(28.8)}} & 
            \makecell{0.81 \\ {\scriptsize(24.6)}} & 
            \makecell{0.96 \\ {\scriptsize(28.9)}} & 
            \makecell{0.86 \\ {\scriptsize(28.0)}} & 
            \makecell{0.83 \\ {\scriptsize(26.2)}} & 
            \makecell{0.94 \\ {\scriptsize(32.7)}} &
            \makecell{0.91 \\ {\scriptsize(30.4)}} &
            \makecell{0.95 \\ {\scriptsize(26.2)}} &
            \makecell{0.91 \\ {\scriptsize(28.7)}} \\ 
            
            \midrule 
            \multirow{6}{*}{TX} &  MB & 
        \makecell{0.96 \\
        {\scriptsize(30.3)}} & \makecell{1.01 \\
        {\scriptsize(33.0)}} & \makecell{0.98 \\
        {\scriptsize(33.2)}} & \makecell{0.98 \\
        {\scriptsize(31.5)}} & \makecell{1.01 \\
        {\scriptsize(31.6)}} & \makecell{0.90 \\
        {\scriptsize(28.4)}} & \makecell{0.97 \\
        {\scriptsize(31.4)}} & \makecell{0.98 \\
        {\scriptsize(25.8)}} & \makecell{0.72 \\
        {\scriptsize(26.7)}} & \makecell{1.06 \\
        {\scriptsize(35.7)}} & \makecell{0.90 \\
        {\scriptsize(28.4)}} & \makecell{0.81 \\
        {\scriptsize(25.9)}} & \makecell{0.83 \\
        {\scriptsize(27.0)}} \\
        \cmidrule{2-15}
       &chrF &
            \makecell{0.95 \\ {\scriptsize(48.2)}} &
            \makecell{1.00 \\ {\scriptsize(52.3)}} & 
            \makecell{0.98 \\ {\scriptsize(46.6)}} & 
            \makecell{0.99 \\ {\scriptsize(51.0)}} &
            \makecell{1.01 \\ {\scriptsize(52.4)}} & 
            \makecell{0.93 \\ {\scriptsize(46.5)}} & 
            \makecell{0.97 \\ {\scriptsize(50.9)}} & 
            \makecell{0.80 \\ {\scriptsize(30.7)}} & 
            \makecell{0.88 \\ {\scriptsize(49.1)}} & 
            \makecell{1.00 \\ {\scriptsize(52.6)}} &
            \makecell{0.93 \\ {\scriptsize(48.9)}} &
            \makecell{0.87 \\ {\scriptsize(46.2)}} &
            \makecell{0.92 \\ {\scriptsize(46.2)}} \\    
            \cmidrule{2-15}        
        &SB &
            \makecell{0.98 \\ {\scriptsize(35.3)}} &
            \makecell{1.02 \\ {\scriptsize(34.0)}} & 
            \makecell{1.03 \\ {\scriptsize(33.2)}} & 
            \makecell{0.98 \\ {\scriptsize(32.9)}} &
            \makecell{1.02 \\ {\scriptsize(33.7)}} & 
            \makecell{0.92 \\ {\scriptsize(29.2)}} & 
            \makecell{0.97 \\ {\scriptsize(33.8)}} & 
            \makecell{0.90 \\ {\scriptsize(26.9)}} & 
            \makecell{0.75 \\ {\scriptsize(28.3)}} & 
            \makecell{0.99 \\ {\scriptsize(35.4)}} &
            \makecell{0.93 \\ {\scriptsize(31.1)}} &
            \makecell{0.89 \\ {\scriptsize(36.8)}} &
            \makecell{0.86 \\ {\scriptsize(30.2)}} \\     
            \midrule
            \\
            CI Err. & (E-3) &
         
            \makecell{4.5 \\ {\scriptsize(0.17)}} & 
            \makecell{1.5 \\ {\scriptsize(0.13)}} & 
            \makecell{2.7 \\ {\scriptsize(0.11)}} &
            \makecell{2.6 \\ {\scriptsize(0.17)}} & 
            \makecell{2.4 \\ {\scriptsize(0.15)}} & 
            \makecell{1.8 \\ {\scriptsize(0.12)}} & 
            \makecell{1.7 \\ {\scriptsize(0.23)}} & 
            \makecell{5.7 \\ {\scriptsize(0.30)}} & 
            \makecell{3.0 \\ {\scriptsize(0.23)}} &
            \makecell{2.2 \\ {\scriptsize(0.11)}} &
            \makecell{2.2 \\ {\scriptsize(0.16)}} &
            \makecell{2.5 \\ {\scriptsize(0.24)}}  & 
            \makecell{3.1 \\ {\scriptsize(0.22)}} \\  
         \bottomrule \end{tabular} \caption
    {\bleu score ratios between pairs of noisy and normalized
      sets of sentences, containing only one UGC
      specificity. \bleu scores on noisy sets are shown in
      parenthesis. {\em Three different metrics are shown for comparison: \texttt{MultiBleu-detok.perl} (MB) , \texttt{chrF}
      and \texttt{SacreBleu} (SB). {\em Error for 95\% confidence intervals (CI Err).}
      } \label{tab:detailled_BLEU_ratio_1error}}
}
\end{table*}
\FloatBarrier

\begin{figure*}[!ht]
    \centering
    \resizebox{0.36\columnwidth}{!}{

    \begin{tikzpicture}
\begin{axis}[
	xlabel=Number of present UGC specificities per sent.,
	ylabel=Ratio,
	width=7cm,height=6cm,
    legend style={at={(0.7,.8)},anchor=west},
    xtick={1,2,3,4.34},
    xticklabels={$1$,$2$,$3$,$4to7$}
    ]
\addplot[color=red,mark=x] coordinates {
	(1, 0.9)
	(2, 0.83)
	(3, 0.77)
	(4.34, 0.75)
};

\addplot[color=blue,mark=*] coordinates {
	(1, 0.92)
	(2, 0.87)
	(3, 0.83)
	(4.34, 0.83)

};

\addplot[color=green,mark=o] coordinates {
	(1, 0.96)
	(2, 0.89)
	(3, 0.86)
	(4.34, 0.84)

};

\legend{s2s,c2c,tx}
\end{axis}
\end{tikzpicture}
    }
    
\caption{Noisy/Clean \bleu scores' ratios for accumulated number of UGC specificities present per sentence for each model, corresponding to the results in Table~\ref{tab:BLEU_ratio_Nerror}. The \textit{4to7} bin groups more than 4 types to provide a larger subcorpus, which weighted average is 4.34 UGC specificities per sentence. \label{fig:sub1}}
\end{figure*}
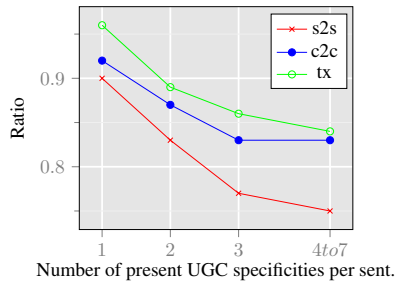

\newpage


\pagebreak

\section{Reproducibility}
\paragraph*{Data}
All the UGC test sets and source code for our experiments are provided in the supplementary materials. For training data, we let the reader refer to each project's website for \texttt{WMT}\footnote{\url{https://www.statmt.org/wmt15/translation-task.html}} (consisting of \texttt{Europarlv7}, \texttt{Newcommentariesv10} and \texttt{Open Subtitles}\footnote{\url{http://opus.nlpl.eu/download.php?f=OpenSubtitles/v2018/moses/en-fr.txt.zip}}, both accessed on November, 2019. Regarding clean test sets, we used \texttt{newstest15} from \texttt{WMT} and a subset of 11,000 unique phrases from \texttt{Open Subtitles}. We make the former test available in the link provided above for exact performance comparison.

\paragraph*{Computation}
The NMT systems were trained using 1 Tesla V100, during an average of 72 hours to converge to the final solution for the \texttt{char2char} model and 56 hours for the BPE-based baselines. 
\subsection{NMT Models}
\paragraph*{Character-based models}

\texttt{char2char} models were trained as
out-of-the box systems using the implementations provided by \cite{DBLP:journals/tacl/LeeCH17_et_al} .\footnote{\url{https://github.com/nyu-dl/dl4mt-c2c}}

\paragraph*{BPE-based models}

We consider, as baseline, two standard NMT models that take, as
input, tokenized sentences. The first one is a \texttt{seq2seq}
bi-LSTM architecture with global attention decoding. The
\texttt{seq2seq} model was trained using the XNMT
toolkit~\cite{DBLP:conf/amta/NeubigSWFMPQSAG18}.\footnote{We decided
  to use XNMT, instead of OpenNMT in our experiments in order to
  compare our results to the ones of \newcite{michel2018mtnt}.} It
consists in a 2-layered Bi-LSTM layers encoder and 2-layered Bi-LSTM
decoder. It considers, as input, word embeddings of 512~components and
each LSTM units has 1,024 components. 

We also study a vanilla Transformer model using the implementation
proposed in the OpenNMT framework
\cite{DBLP:conf/amta/KleinKDNSR18}. It consists of 6 layers with word
embeddings of 512 components, a feed-forward layers made of 2,048
units and 8 self-attention heads.


\paragraph*{Hyper-parameters}
In Table~\ref{tab:hyperparams}, we list the training variables set for our experiments. They match their corresponding default hyper-parameters.
\begin{table}[!htpb]
\centering
\begin{tabular}{l|l}
\toprule
   Batch size  & 64 \\
    Optimizer & Adam \\
    Learning rate & 1e-4 \\
    Epochs    &   10 (best of) \\
    Patience   &  2 epochs \\
    Gradient clip & [-1.0, 1.0] \\
    \bottomrule
\end{tabular}
\caption{Hyper-parameters used for training the NMT systems. \label{tab:hyperparams}}
\end{table}


\paragraph*{Pre-processing}
For the BPE models, we used a 16K merging operations tokenization employing \texttt{sentencepiece}\footnote{\url{https://github.com/google/sentencepiece}}. For word-level statistics we segmented the corpora using the \texttt{Moses} tokenizer \footnote{\url{https://github.com/moses-smt/mosesdecoder}}.

\end{document}